\pdfoutput=1

\documentclass[11pt,dvipsnames]{article}

\usepackage[]{formatting/acl2022/acl}

\usepackage{times}
\usepackage{latexsym}

\usepackage[T1]{fontenc}

\usepackage[utf8]{inputenc}

\usepackage{microtype}

\usepackage{nccmath}
\usepackage{graphicx}
\graphicspath{{figures/}}
\DeclareGraphicsExtensions{.png,.pdf}

\usepackage[inline,shortlabels]{enumitem}

\usepackage{multirow}
\usepackage{ctable} 

\usepackage{amsmath}  
\usepackage{amsfonts}
\usepackage[scaled=0.86]{helvet}  
\usepackage{kbordermatrix}

\usepackage{flushend}

\newcommand{\secref}[1]{\S\ref{#1}} 

\newcommand{\figref}[1]{Fig.~\ref{#1}}    
\newcommand{\tabref}[1]{Table~\ref{#1}}  
\newcommand{\Tabref}[1]{Table~\ref{#1}}  
\newcommand{\equref}[1]{eq.~(\ref{#1})}
\newcommand{\eqsref}[2]{eq.~(\ref{#1})--(\ref{#2})}

\newcommand{\eg}{e.g., }

\newcommand{\ie}{i.e., }

\newcommand{\dwiedataset}{DWIE}

\newcommand{\originalaida}{AIDA}
\newcommand{\ouraida}{AIDA$^{+}$}
\newcommand{\ouraidaseta}{AIDA$^{+}_\textrm{a}$}
\newcommand{\ouraidasetb}{AIDA$^{+}_\textrm{b}$}
\newcommand{\clinker}{{\textsf{Local}}}
\newcommand{\mtt}{{\textsf{Global}}}

\newcommand{\Baseline}{\textsf{Standalone}}

\newcommand{\CEAFe}{CEAF$_\textrm{e}$}

\definecolor{myRed}{RGB}{204,0,0}
\definecolor{myGreen}{RGB}{0,84,0}

\title{Towards Consistent Document-level Entity Linking:\\Joint Models for Entity Linking and Coreference Resolution}

\author{Klim Zaporojets, Johannes Deleu, Yiwei Jiang, Thomas Demeester, Chris Develder \\ 
  Ghent University – imec, IDLab \\
  Ghent, Belgium \\
\texttt{\{first\_name.last\_name\}@ugent.be} \\}

\begin{document}
\maketitle
\begin{abstract}
We consider the task of 
document-level 
entity linking (EL), where it is important to make consistent decisions for entity mentions over the full document jointly.
We aim to leverage explicit ``connections'' among mentions within the document itself:~we propose to join EL and coreference resolution (coref) in a \textit{single} structured prediction task over directed trees and use a globally normalized model to solve it. This contrasts with related works where two separate models are trained for each of the tasks and additional logic is required to merge the outputs.~Experimental results 
on two datasets show a boost of up to +5\% F1-score on both coref and EL tasks, compared to their standalone counterparts.
For a subset of hard cases, with individual mentions lacking the correct EL in their candidate entity list, we obtain a 
+50\%
increase in 
accuracy.\footnote{Our code, models and \ouraida~dataset will be released on \url{https://github.com/klimzaporojets/consistent-EL} }
\end{abstract}

\section{Introduction}
\label{sec:intro}
In this paper we explore a principled approach to solve entity linking (EL) jointly with coreference resolution (coref). Concretely, we formulate coref+EL
as a \textit{single} structured task over directed trees that conceives EL and coref as two complementary components: a coreferenced cluster can only be linked to a single entity or NIL (\ie a non-linkable entity), and all mentions linking to the same entity are coreferent. This contrasts with previous attempts to join coref+EL \citep{hajishirzi2013joint, dutta2015c3el, angell2021clustering} where coref and EL models are trained separately and additional logic is required to merge the predictions of both tasks. 
\begin{figure}[t]
\centering
\includegraphics[width=.95\columnwidth]{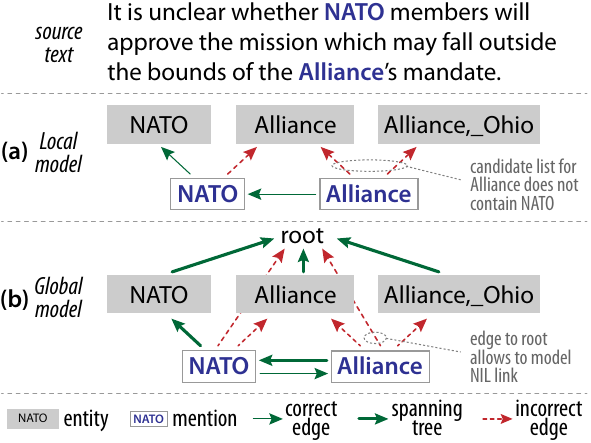}
\captionsetup{singlelinecheck=off}
\caption[test]{Illustration of our 2 explored graph models:
\begin{enumerate*} [(a)]
    \item {\clinker} where edges are only allowed from spans to antecedents or candidate entities, and
    \item {\mtt} where the prediction involves a spanning tree over all nodes.  
\end{enumerate*}}
\label{fig:architectures}
\end{figure}

Our first approach ({\clinker} in \figref{fig:architectures}(a)) is motivated by current state-of-the-art 
coreference resolution models 
\citep{joshi2019bert,wu2020corefqa}
that predict a single antecedent for each span to resolve.
We extend this architecture by also considering entity links as potential antecendents:
in the example of \figref{fig:architectures}, the mention ``Alliance'' can be either connected to its antecedent mention ``NATO'' or to any of its candidate links (\emph{Alliance} or \emph{Alliance,\_Ohio}). 
While straightforward,
this approach cannot solve 
cases where the first coreferenced mention does not include the correct entity in its candidate list
(\eg if the order of ``NATO'' and ``Alliance'' mentions in \figref{fig:architectures} would be reversed). 
We therefor propose a second approach, {\mtt}, which by construction overcomes this inherent limitation by using bidirectional connections between mentions. 
Because that implies cycles could be formed, we resort to solving a maximum spanning tree problem. 
Mentions that refer to the same entity form a cluster, represented as a subtree rooted by the single entity they link to. 
To encode the overall document's clusters in a single spanning tree, we introduce a virtual \textit{root} node
(see \figref{fig:architectures}(b)).\footnote{Coreference clusters without a linked entity, \ie a NIL cluster, have a link of a mention directly to the root.}

This paper contributes: \begin{enumerate*}[(i)]
    \item 2 architectures ({\clinker} and {\mtt}) for joint entity linking (EL) and corefence resolution, 
    \item an extended AIDA dataset \citep{hoffart2011robust}, adding new annotations of linked and NIL coreference clusters,
    \item experimental analysis on 2 datasets where our joint coref+EL models achieve up to +5\% F1-score on both tasks compared to standalone models. We also show up to 
    +50\% in accuracy
    for hard cases of EL 
    where
    entity mentions 
    lack
    the correct entity in their candidate list. 
\end{enumerate*}

\section{Architecture}
\label{sec:architecture}

Our model takes as input
\begin{enumerate*}[(i)]
    \item the full document text, and
    \item an \emph{alias table} with entity candidates for each of the possible spans. 
\end{enumerate*} 
Our end-to-end approach allows to jointly predict the mentions, entity links and coreference relations between them. 

\subsection{Span and Entity Representations}

We use SpanBERT (base) from \citet{joshi2020spanbert} to obtain \emph{span} representations $\textbf{g}_i$ for a particular span $s_i$. 
Similarly to \citet{luan2019general,xu2020revealing}, we apply an additional pruning step to keep only the top-$N$ spans based on the pruning score $\Phi_{\mathrm{p}}$ from a feed-forward neural net (FFNN): 
\begin{equation}
\Phi_{\mathrm{p}}(s_i) = \mathrm{FFNN}_{P}( \textbf{g}_i).
\end{equation}

For a candidate entity $e_{j}$ of span $s_i$ we 
will obtain
representation as $\textbf{e}_{j}$ 
(which is further detailed in \secref{sec:experimental_setup}). 

\subsection{Joint Approaches}
\label{sec:joint_approaches}
We propose two 
methods for joint coreference and EL.
The first, {\clinker}, is motivated by end-to-end span-based coreference resolution models 
\citep{lee2017end,lee2018higher} 
that optimize the marginalized probability of the correct antecedents for each given span.
We extend this local marginalization to include the span's candidate entity links. Formally, the 
modeled 
probability of 
$y$ 
(text span or candidate entity) 
being the antecedent of span $s_i$ is: 
\begin{equation}
P_{\mathrm{cl}}(y | s_i) = \dfrac{\exp\big(\Phi_{\mathrm{cl}}(s_i,y)\big)}{\sum_{y'\in \mathcal{Y}(s_i)} \exp \big( \Phi_{\mathrm{cl}}(s_i,y') \big) }, \label{eq:softmax}
\end{equation}
where $\mathcal{Y}(s_i)$ is the set of antecedent spans unified with the candidate entities for $s_i$. For antecedent \emph{spans} $\{s_j: j < i\}$ the score $\Phi_{\mathrm{cl}}$ is defined as:
\begin{align}
\medmath{\Phi_{\mathrm{cl}}(s_i, s_j) = \Phi_{\mathrm{p}}(s_i) + \Phi_{\mathrm{p}}(s_j) + \Phi_{\mathrm{c}}(s_i, s_j)}, \label{eq:span_cl1} \\ 
\medmath{\Phi_{\mathrm{c}}(s_i, s_j) = \mathrm{FFNN}_{C}([ \textbf{g}_i; \textbf{g}_{j}; \textbf{g}_{i} \odot \textbf{g}_{j}; \boldsymbol{\varphi}_{i,j}])}, \label{eq:span_cl2}
\end{align}
where $\boldsymbol{\varphi}_{i,j}$ is an embedding encoding the distance\footnote{Measured in number of spans, after pruning.} between spans $s_i$ and $s_j$.
Similarly, for a particular candidate \emph{entity} $e_{j}$, the score $\Phi_{\mathrm{cl}}$ is:
\begin{align}
    \Phi_{\mathrm{cl}}(s_i, e_{j}) = \Phi_{\mathrm{p}}(s_i) + \Phi_{{\ell}}(s_i, e_{j}), \label{eq:entity_cl1} \\
    \Phi_{\ell}(s_i, e_{j}) = \mathrm{FFNN}_{L}([\textbf{g}_i; \textbf{e}_{j}]). \label{eq:entity_cl2}
\end{align}
An example graph of mentions and entities with edges for which aforementioned scores $\Phi_{\mathrm{cl}}$ would be calculated is sketched in \figref{fig:architectures}(a). 
While simple, this approach fails
to correctly solve EL when the correct entity is only present in the candidate lists of mention spans occurring later in the text (since 
earlier
mentions have no access to it).

To solve EL in the general case, even
when the first mention does not have the correct entity, we propose bidirectional connections between mentions, thus leading to a maximum spanning tree problem in our {\mtt} approach.
Here we define a score for a (sub)tree $t$, noted as $\Phi_\mathrm{tr}(t)$:
\begin{equation}
   \Phi_\mathrm{tr}(t) = \sum_{(i,j) \in t} \Phi_{\mathrm{cl}}(u_i, u_j), \label{eq:mtt1}   
\end{equation}
where $u_i$ and $u_j$ are two connected nodes (\ie \emph{root}, candidate entities or spans) in $t$.
For a ground truth cluster $c \in C$ (with $C$ being the set of all such clusters), with its set\footnote{For a single cluster annotation, indeed it is possible that multiple correct trees can be drawn.} of correct subtree representations $\mathcal{T}_c$, we model the cluster's likelihood with its subtree scores. We minimize the negative log-likelihood $\mathcal{L}$ of all clusters:
\begin{align}
\mathcal{L} &= - \log \frac{\prod_{c \in C} \sum_{t \in \mathcal{T}_c} \exp \big(\Phi_\mathrm{tr}(t)\big)}{\sum_{t \in \mathcal{T}_\textit{all}} \exp \big( \Phi_\mathrm{tr}(t) \big)}. \label{eq:loss_naive}
\end{align}%
Naively enumerating all possible spanning trees ($\mathcal{T}_\textit{all}$ or $\mathcal{T}_c$) implied by this equation is infeasible, since their number is exponentially large.
We use the adapted Kirchhoff's Matrix Tree Theorem 
(MTT; \citet{koo2007structured,william1984tutte})
to solve this:
the sum of the weights of the spanning trees in a directed graph rooted in \emph{r} is equal to the determinant of the Laplacian matrix of the graph with the row and column corresponding to \emph{r} removed (\ie the \emph{minor} of the Laplacian with respect to \emph{r}). This way, \equref{eq:loss_naive} can be rewritten as
\begin{align}
    \mathcal{L} &= - \log \frac{\prod_{c \in C} {\det \Big(\mathbf{\hat{L}}_{c} \big(\mathbf{\Phi_\mathrm{cl}}\big) \Big)}}{\det \Big(\mathbf{L}_{r}\big(\mathbf{\Phi_\mathrm{cl}}\big)\Big)}, \label{eq:loss_mtt}
\end{align}
where $\mathbf{\Phi_\mathrm{cl}}$ is the weighted adjacency matrix of the graph, and $\mathbf{L}_{r}$ is the minor of the Laplacian with respect to the root node $r$. An entry in the Laplacian matrix is 
calculated
as
\begin{align}
    \medmath{L_{i,j} = 
    \begin{cases}
      \sum\limits_{k} \exp(\Phi_{\mathrm{cl}}(u_k, u_j)) & \text{if $i = j$}\\
      - \exp(\Phi_{\mathrm{cl}}(u_i, u_j)) & \text{otherwise}
    \end{cases}}, \label{eq:laplacian}
\end{align}
Similarly, $\mathbf{\hat{L}}_{c}$ is a \textit{modified Laplacian} matrix where the first row is replaced with the root $r$ selection scores $\Phi_{\mathrm{cl}}(r, u_j)$. 
For clarity, Appendix~\ref{sec:appendix} presents a toy example with detailed steps to calculate the loss in \equref{eq:loss_mtt}.

To calculate the scores of each of the entries $\Phi_\textrm{cl} (u_i, u_j)$ to $\mathbf{\Phi_\mathrm{cl}}$ matrix in 
eqs.~(\ref{eq:mtt1}) and~(\ref{eq:loss_mtt})
for {\mtt}, we use the same approach as in {\clinker} for edges between two mention spans, or between a mention and entity. 
For the directed edges between the root $r$ and a candidate entity $e_j$ we choose $\Phi_{\mathrm{cl}}(r, e_j)=0$.
Since we represent NIL clusters by edges from the 
mention spans directly to the root,
we also need scores for them: we use \equref{eq:span_cl1} with $\Phi_\mathrm{p}(r)=0$.
We use 
Edmonds'
algorithm \citep{edmonds1967optimum} for decoding the maximum spanning tree.  
\section{Experimental Setup}
\label{sec:experimental_setup}

We considered two datasets to evaluate our proposed models: {\dwiedataset} \citep{zaporojets2021dwie} and {\originalaida} \citep{hoffart2011robust}.
Since {\originalaida} essentially does not contain coreference information, we had to extend it by
\begin{enumerate*}[(i)]
    \item adding missing mention links in order to make annotations consistent on the coreference cluster level, and 
    \item annotating NIL coreference clusters.
\end{enumerate*}
We note this extended dataset as {\ouraida}. See \tabref{tab:dataset} for the details. 

As input to our models, for {\dwiedataset} we generate spans of up to 5 tokens.
For each mention span $s_i$, we find candidates from a dictionary of entity surface forms used for hyperlinks in Wikipedia.
We then keep the top-16 candidates based on the prior for that surface form, as per \citet[\S3]{yamada2016joint}.
Each of those candidates $e_{j}$ is represented using a Wikipedia2Vec embedding $\textbf{e}_{j}$ \citep{yamada2016joint}.\footnote{We use Wikipedia version 20200701.} 
For {\ouraida}, we use 
the spans, entity candidates, and entity representations
from \citet{kolitsas2018end}.\footnote{\url{https://github.com/dalab/end2end_neural_el}}
\begin{table}[t]
    \centering
    \resizebox{1.0\columnwidth}{!}
    {\begin{tabular}{lcccc}
        \toprule
         \multirow{2}{*}{Dataset} & \# Linked & \# NIL & Linked & \# NIL \\
          & clusters & clusters & mentions & mentions \\         
         \midrule
	 \dwiedataset & 11,967 & 9,935 & 28,482 & 14,891 \\ 
	 \originalaida & 16,673 & - & 27,817 & 7,112 \\ 
	 \ouraida & 16,775 & 4,284 & 28,813 & 6,116 \\     \bottomrule
    \end{tabular}}
    \caption{Datasets statistics.} 
    \label{tab:dataset}
\end{table}

To assess the performance of our joint coref+EL models {\clinker} and {\mtt}, we also provide 
\Baseline~implementations for coref and EL tasks. The \Baseline~coref model is trained using only the coreference component of our joint architecture (\eqsref{eq:softmax}{eq:span_cl2}), while the EL model is based only on the linking component (\equref{eq:entity_cl2}).

As performance metrics, for coreference resolution we calculate the average-F1 score of commonly used MUC~\citep{vilain1995model}, B$^3$~\citep{bagga1998algorithms} and {\CEAFe} \citep{luo2005coreference} metrics as implemented by \citet{pradhan2014scoring}.
For EL, we use
\begin{enumerate*}[(i)]
    \item \emph{mention}-level F1 score (EL$_\mathrm{m}$), and
    \item \emph{cluster}-level \textit{hard} F1 score (EL$_\mathrm{h}$) that counts a true positive only if both the coreference cluster (in terms of all its mention spans) and the entity link are correctly predicted. 
\end{enumerate*}
These EL metrics are executed in a \emph{strong matching} setting that requires predicted spans to exactly match the boundaries of gold mentions.
Furthermore, for EL we only report the performance on non-NIL 
mentions, leaving the study of NIL links for future work. 

Our experiments will answer the following research questions:~\begin{enumerate*}[label=\textbf{(Q\arabic*)}]
   \item \label{it:q-performance} How does performance of our joint coref+EL models compare to \Baseline~
   models?
    \item \label{it:q-coherence} Does jointly solving coreference resolution and EL enable more coherent EL predictions?
    \item \label{it:q-corner} How do our joint models perform on hard cases where some individual entity mentions do not have the correct candidate?
\end{enumerate*}

\section{Results}
\label{sec:results}
\begin{table*}[t]
\centering
\resizebox{0.95\textwidth}{!}{
\begin{tabular}{c ccc c ccc c ccc}
\toprule
& \multicolumn{3}{c}{\dwiedataset} &&  \multicolumn{3}{c}{\ouraidaseta} && \multicolumn{3}{c}{\ouraidasetb} \\ 
\cmidrule(lr){2-4}\cmidrule(lr){6-8}\cmidrule(lr){10-12} 
Setup & EL$_\mathrm{m}$ & EL$_\mathrm{h}$ & Coref && EL$_\mathrm{m}$ & EL$_\mathrm{h}$ & Coref && EL$_\mathrm{m}$ & EL$_\mathrm{h}$ & Coref \\ 
         \midrule
         \Baseline & 88.7${\scriptstyle \pm\text{0.1}}$ & 78.4${\scriptstyle \pm\text{0.2}}$ & 94.5${\scriptstyle \pm\text{0.1}}$ && 86.2${\scriptstyle \pm\text{0.4}}$ & 80.7${\scriptstyle \pm\text{0.5}}$ & 93.8${\scriptstyle \pm\text{0.1}}$ && 79.1${\scriptstyle \pm\text{0.3}}$ & 74.0${\scriptstyle \pm\text{0.3}}$ & 91.5${\scriptstyle \pm\text{0.3}}$ \\
         \clinker & 90.5${\scriptstyle \pm\text{0.4}}$ & 83.4${\scriptstyle \pm\text{0.4}}$ & 94.4${\scriptstyle \pm\text{0.2}}$ && 87.5${\scriptstyle \pm\text{0.2}}$ & 83.1${\scriptstyle \pm\text{0.2}}$ & 94.7${\scriptstyle \pm\text{0.1}}$ && \textbf{79.9}${\scriptstyle \pm\textbf{0.4}}$ & 75.8${\scriptstyle \pm\text{0.3}}$ & \textbf{92.3${\scriptstyle \pm\text{0.1}}$} \\
         \mtt & \textbf{90.7}${\scriptstyle \pm\textbf{0.3}}$ & \textbf{83.9}${\scriptstyle \pm\text{0.5}}$ & \textbf{94.7}${\scriptstyle \pm\text{0.2}}$ && \textbf{87.6}${\scriptstyle \pm\textbf{0.2}}$ & \textbf{83.7}${\scriptstyle \pm\text{0.3}}$ & \textbf{95.1}${\scriptstyle \pm\text{0.1}}$ && 79.6${\scriptstyle \pm\text{0.4}}$ & \textbf{76.0}${\scriptstyle \pm\text{0.4}}$ & 92.2${\scriptstyle \pm\text{0.2}}$ \\ 
\bottomrule 
	\end{tabular}
}
	\caption{Experimental results 
	(F1 scores defined in \secref{sec:experimental_setup}) 
	using the 
	\Baseline~coreference and EL
	models compared to our joint architectures (\clinker~and \mtt), on \dwiedataset~and \ouraida~datasets.}
	\label{tab:overview_results}
\end{table*}
\Tabref{tab:overview_results} shows the results of our compared models for EL and 
coreference resolution 
tasks.
Answering \ref{it:q-performance}, we observe a general improvement in performance of our coref+EL joint models ({\clinker} and {\mtt}) compared to \Baseline~
on the EL task.
Furthermore, this difference is bigger when using our cluster-level \textit{hard} metrics.
This also answers \ref{it:q-coherence} by indicating that the joint models tend to produce more coherent cluster-based predictions.
To make this more explicit, \Tabref{tab:cluster_coherence} compares the accuracy for singleton clusters (\ie clusters composed by a single entity mention), denoted as $S$, to that of clusters composed by multiple mentions, denoted as $M$.
We observe that the difference in performance between our joint models and
\Baseline~is bigger on $M$ clusters (with a consistent 
superiority 
of \mtt), indicating that our approach indeed produces more coherent predictions for mentions that refer to the same concept. 
\begin{table}[t]
    \centering
    \resizebox{1.0\columnwidth}{!}{
    \begin{tabular}{l cc c cc c cc}
        \toprule
         & \multicolumn{2}{c}{\dwiedataset} && \multicolumn{2}{c}{\ouraidaseta} && \multicolumn{2}{c}{\ouraidasetb}  \\
         \cmidrule(lr){2-3}\cmidrule(lr){5-6}\cmidrule(lr){8-9} 
         Setup & $S$ & $M$ && $S$ & $M$ && $S$ & $M$ \\
         \midrule
         \Baseline & 80.4 & 69.5 && 82.9 & 70.7 && 77.0 & 57.0 \\
         \clinker & \textbf{82.6} & 78.6 && 84.9 & 74.8 && \textbf{79.8} & 61.4 \\
         \mtt & \textbf{82.6} & \textbf{80.0} && \textbf{85.1} & \textbf{76.8} && 79.3 & \textbf{63.0} \\                  
\bottomrule
    \end{tabular}
    } 
    \caption{Cluster-based accuracy of link prediction on singletons ($S$) and clusters of multiple mentions ($M$).} 
    \label{tab:cluster_coherence}
\end{table}
Further analysis reveals that this difference in performance is even higher for a more complex scenario where the clusters contain mentions with different surface forms (not 
shown 
in the table). 
\begin{table}[t]
    \centering
    \small
    \begin{tabular}{lccc}
        \toprule
         {Setup} & \dwiedataset & \ouraidaseta & \ouraidasetb  \\
         \midrule
	 \Baseline & 0.0 & 0.0 & 0.0  \\ 
	 \clinker & 41.7 & 27.4 & 26.9  \\ 
	 \mtt & \textbf{57.6} & \textbf{50.2} & \textbf{29.7} \\     \bottomrule
    \end{tabular}
    \caption{EL accuracy for corner case mentions where the correct entity is not in the mention's candidate list.} 
    \label{tab:results_corner}
\end{table}

In order to tackle research question \ref{it:q-corner}, we study the accuracy of our models on the important corner case that involves mentions without correct entity in their candidate lists.
This is illustrated in \Tabref{tab:results_corner}, which focuses on such mentions 
in clusters where at least one mention contains the correct entity in its candidate list.
As expected, the 
\Baseline~model cannot link such mentions, as it is limited to the local 
candidate list.
In contrast, both our joint approaches can solve some of these cases by using the correct candidates from other mentions in the cluster, with a superior performance of our {\mtt} model compared to the {\clinker} one.

\section{Related Work}
\label{sec:related}
\textbf{Entity Linking:} Related work in entity linking (EL) tackles the document-level linking coherence by exploring relations between entities \citep{kolitsas2018end,yang2019learning,le2019boosting}, or entities and mentions \citep{le2018improving}.
More recently, contextual BERT-driven \citep{devlin2019bert} language models have been used for the EL task  \citep{broscheit2019investigating,de2020autoregressive,de2021highly,yamada2020global} by jointly embedding mentions and entities. 
In contrast, we explore a cluster-based EL approach where the coherence is achieved on \textit{coreferent} entity mentions level.

\noindent\textbf{Coreference Resolution:} Span-based antecedent-ranking coreference resolution \citep{lee2017end,lee2018higher} has 
seen
a recent boost by using SpanBERT representations \citep{xu2020revealing,joshi2020spanbert,wu2020corefqa}.
We extend this approach in our {\clinker} joint coref+EL architecture.
Furthermore, 
we rely on Kirchhoff's Matrix Tree Theorem \citep{koo2007structured,william1984tutte} to efficiently train a more expressive spanning tree-based {\mtt} method. 

\noindent\textbf{Joint EL+Coref:} \citet{fahrni2012jointly} 
introduce a more expensive rule-based Integer Linear Programming component to jointly predict coref and EL. 
\citet{durrett2014joint} jointly train coreference and entity linking without enforcing single-entity per cluster consistency. 
More recently, \citet{angell2021clustering,agarwal2021entity} use additional logic to achieve consistent cluster-level entity linking.
In contrast, our 
proposed
approach constrains the space of the predicted spanning trees on a structural level (see \figref{fig:architectures}). 
\section{Conclusion}
\label{sec:conclusion}
We propose two end-to-end models to solve entity linking and coreference resolution tasks in a joint setting. 
Our joint architectures achieve superior performance compared to the standalone counterparts. 
Further analysis reveals that this boost in performance is driven by more coherent predictions on the level of mention clusters (linking to the same entity) and extended candidate entity coverage. 
\bibliography{manuscript_acl}
\bibliographystyle{formatting/acl2022/acl_natbib}
\appendix

\section{Step by Step Example of MTT Theorem}
\label{sec:appendix}
\begin{figure}[t]
\centering
\includegraphics[width=.70\columnwidth,trim={0.5cm 22.5cm 15.5cm 1cm},clip]{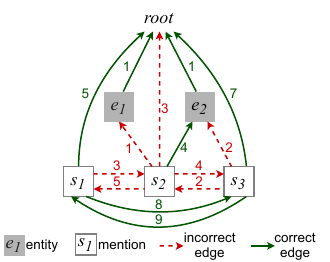}
\captionsetup{singlelinecheck=off}
\caption[test]{Illustrative graph example of {\mtt} model.~The weights of the edges correspond to $\exp(\mathbf{\Phi_{\mathrm{cl}}})$ (see \equref{eq:ex_exp_weight_denominator}). }
\label{fig:illustrative_example}
\end{figure}

In this appendix we will provide a clarifying artificial example 
in order to walk the reader step by step through MTT (\eqsref{eq:loss_mtt}{eq:laplacian}) applied in our \mtt~approach.
The graph of the example is illustrated in \figref{fig:illustrative_example} and is composed by nodes representing $root$ ($r$), entities $e_1$ and $e_2$, and spans $s_1$, $s_2$ and $s_3$. 
The span $s_2$ is associated with candidate entity set $\{e_1, e_2\}$ (\ie represented by edges from $s_2$ to $e_1$ and $e_2$), and $s_3$ with $\{e_2\}$ (\ie represented by the edge from $s_3$ to $e_2$). The candidate entity set of $s_1$ is empty.  
The nodes are grouped in two ground truth clusters: NIL cluster $c_1 = \{s_1, s_2\}$, and linked cluster $c_2 = \{e_2, s_2\}$. 

\renewcommand{\kbldelim}{[}
\renewcommand{\kbrdelim}{]}
The exponential of weighted adjacency matrix\footnote{For simplicity, the weights are small integers.} $\mathbf{\Phi_\mathrm{cl}}$ of the presented example is:
\begin{equation}
\small 
  \exp(\mathbf{\Phi_{\mathrm{cl}}}) = \kbordermatrix{
    & r & e_1 & e_2 & s_1 & s_2 & s_3 \\
    r & 0 & \textcolor{myGreen}{1} & \textcolor{myGreen}{1} & \textcolor{myGreen}{5} & \textcolor{myRed}{3} & \textcolor{myGreen}{7} \\
    e_1 & 0 & 0 & 0 & 0 & \textcolor{myRed}{1} & 0 \\
    e_2 & 0 & 0 & 0 & 0 & \textcolor{myGreen}{4} & \textcolor{myRed}{2} \\
    s_1 & 0 & 0 & 0 & 0 & \textcolor{myRed}{5} & \textcolor{myGreen}{9} \\
    s_2 & 0 & 0 & 0 & \textcolor{myRed}{3} & 0 & \textcolor{myRed}{2} \\
    s_3 & 0 & 0 & 0 & \textcolor{myGreen}{8} & \textcolor{myRed}{4} & 0 
  },\label{eq:ex_exp_weight_denominator}
\end{equation}
where the weights of incorrect edges are represented in \textcolor{myRed}{red} (\ie \textcolor{myRed}{red} dashed edges in \figref{fig:illustrative_example}),
the weights of the correct edges in \textcolor{myGreen}{green} (\ie \textcolor{myGreen}{green} edges in \figref{fig:illustrative_example}),
and the weights between disconnected nodes are set to 0. 

In order to compute the \textit{denominator} of the loss function in \equref{eq:loss_mtt},
the Laplacian of the matrix in \equref{eq:ex_exp_weight_denominator} is calculated as described in \equref{eq:laplacian}, and the row and column corresponding to root $r$ removed (\ie the \textit{minor} $\mathbf{L}_{r}$ with respect to the root):
\begin{equation}
    \mathbf{L}_{r} = \kbordermatrix{
     & e_1 & e_2 & s_1 & s_2 & s_3 \\
    e_1  & 1 & 0 & 0 & -1 & 0 \\
    e_2 & 0 & 1 & 0 & -4 & -2 \\
    s_1 & 0 & 0 & 16 & -5 & -9 \\
    s_2 & 0 & 0 & -3 & 17 & -2 \\
    s_3 & 0 & 0 & -8 & -4 & 20 
  }.
\end{equation}
Following Kirchhoff's Matrix Tree Theorem \citep{koo2007structured,william1984tutte}, the determinant of $\mathbf{L}_{r}$ equals to the sum of the weights of all possible spanning trees of the graph represented in \figref{fig:illustrative_example}: 
\begin{equation}
    \det(\mathbf{L}_{r}) = 3600 = \sum_{t \in \mathcal{T}_\textit{all}} \exp \big( \Phi_\mathrm{tr}(t) \big). \label{eq:denom}
\end{equation}

In order to compute the \textit{numerator} of the loss function in \equref{eq:loss_mtt} (\ie the sum of the weights of the spanning trees of ground truth clusters), we first mask out (set to zero) all the weights assigned to incorrect edges:
\begin{equation}
\small
  \exp(\mathbf{\Phi_{\mathrm{cl}}})' = \kbordermatrix{
    & r & e_1 & e_2 & s_1 & s_2 & s_3 \\
    r & 0 & \textcolor{myGreen}{1} & \textcolor{myGreen}{1} & \textcolor{myGreen}{5} & \textcolor{myRed}{0} & \textcolor{myGreen}{7} \\
    e_1 & 0 & 0 & 0 & 0 & \textcolor{myRed}{0} & 0 \\
    e_2 & 0 & 0 & 0 & 0 & \textcolor{myGreen}{4} & \textcolor{myRed}{0} \\
    s_1 & 0 & 0 & 0 & 0 & \textcolor{myRed}{0} & \textcolor{myGreen}{9} \\
    s_2 & 0 & 0 & 0 & \textcolor{myRed}{0} & 0 & \textcolor{myRed}{0} \\
    s_3 & 0 & 0 & 0 & \textcolor{myGreen}{8} & \textcolor{myRed}{0} & 0 
  }
\end{equation}
Next, the \textit{modified Laplacian} (\ie Laplacian with the first row replaced by root $r$ selection weights) $\mathbf{\hat{L}}$ is calculated for both clusters $c_1$ and $c_2$:
\begin{align}
    \mathbf{\hat{L}}_{c_1} = \kbordermatrix{
     & s_1 & s_3 \\
    r & 5 & 7  \\
    s_3 & -8 & 9 
  } \\
    \mathbf{\hat{L}}_{c_2} = \kbordermatrix{
     & e_2 & s_2 \\
    r & 1 & 0  \\
    s_2 & 0 & 4 
  }
\end{align}
The determinants of $\mathbf{\hat{L}}_{c_1}$ and $\mathbf{\hat{L}}_{c_2}$ equal to the sum of the weights of all spanning trees connecting the nodes in clusters $c_1$ and $c_2$ respectively:
\begin{equation}
    \det(\mathbf{\hat{L}}_{c_1}) = 101 = \sum_{t \in \mathcal{T}_{c_1}} \exp \big(\Phi_\mathrm{tr}(t)\big) \label{eq:numer_1}
\end{equation}
\begin{equation}
    \det(\mathbf{\hat{L}}_{c_2}) = 4 = \sum_{t \in \mathcal{T}_{c_2}} \exp \big(\Phi_\mathrm{tr}(t)\big) \label{eq:numer_2}
\end{equation}
Finally, in order to calculate the final loss, we replace the obtained results in eqs.~(\ref{eq:denom}), (\ref{eq:numer_1}), and (\ref{eq:numer_2})
in the loss function of \equref{eq:loss_mtt}: 
\begin{equation}
    \mathcal{L} = - \log \frac{101*4}{3600}. \label{eq:ex_loss}
\end{equation}
\textit{Note}: strictly speaking, there are \textit{three} clusters rooted in \textit{root} in the graph of \figref{fig:illustrative_example}, the third one being $c_3=\{e_1\}$, whose exponential weight is 1 by definition of $\Phi_{\mathrm{cl}}(r, e_j) = 0$ (see \secref{sec:joint_approaches}), and has no impact in calculation of the loss function in \equref{eq:ex_loss}. 
\end{document}